\documentclass[12pt]{iopart}
\usepackage{url}
\usepackage{graphicx}
\usepackage{subfigure}
\usepackage{graphicx}
\usepackage{multirow}
\usepackage[table,xcdraw]{xcolor}

\begin{document}

\title[]{Multi-Class Classification of Blood Cells - \\ End to End Computer Vision based diagnosis case study}

\author{Sai Sukruth Bezugam}

\address{Electrical Engineering Department\\ Indian Institute of Technology Delhi}
\ead{saisukruthbezugam@ieee.org}
\vspace{10pt}
\begin{indented}
\item[]August 2020
\end{indented}

\begin{abstract}

The diagnosis of blood-based diseases often involves identifying and characterizing patient blood samples. Automated methods to detect and classify blood cell subtypes have important medical applications. Automated medical image processing and analysis offers a powerful tool for medical diagnosis. In this work we tackle the problem of white blood cell classification based on the morphological characteristics of their outer contour, color. The work we would explore a set of preprocessing and segmentation (\textbf{Color-based segmentation, Morphological processing, contouring}) algorithms along with a set of features extraction methods ( \textbf{Corner detection algorithms and Histogram of Gradients (HOG)}), dimentionality reduction algorithms (\textbf{Principal Component Analysis (PCA)}) that are able to recognize and classify through various Unsupervised (\textbf{k-nearest neighbors}) and Supervised (\textbf{Support Vector Machine, Decision Trees, Linear Discriminant Analysis, Quadratic Discriminant Analysis, Naïve Bayes}) algorithms different categories of white blood cells to \textit{\textbf{Eosinophil, Lymphocyte, Monocyte, and Neutrophil}}. We even take a step forwards to explore various Deep Convolutional Neural network architecture (\textbf{Sqeezent,  MobilenetV1, MobilenetV2, InceptionNet etc.}) without preprocessing/segmentation and with preprocessing.
We would like to explore many algorithms to identify the robust algorithm with least time complexity and low resource requirement. The outcome of this work can be a cue to selection of algorithms as per requirement for automated blood cell classification. 
\end{abstract}

%
%
%
%
%


\section{Literature survey and report overview}
\subsection{Problem Statement Overview}
Microscopic evaluation of peripheral blood smears has been the
task of hematologists and pathologists. Development of
automated cell counter has transferred this time-consuming job
from human subjects to automated systems. Nevertheless, these
systems have their disadvantages and application restrictions. 
In medical field, with the advancement in technology, the need
for faster and more accurate analysis tools is crucial (e.g. X-ray
machines, CBC machines, MRI etc.), these automated
medical tools are essential for diagnosing patients and their future prognoses of the conditions.
By the advancement of quantitative microscopic techniques, such problems can be overcome by facilitating the Peripheral Blood Smear (PBS) analysis of white blood cells for malignant
disease like leukemia. At the moment, detection of
hematological disorders is done through visual inspection of microscopic images by classifying leukocytes into different types based on some parameters (like color, texture, geometrical, shape features) that lead to early diagnosis of
many diseases.
As per a survey carried out by Indian Association of Blood Cancer and Allied Diseases (IABCD), among all childhood pediatric diseases, leukemia accounts for one-third of childhood cancer in India. In India, use of conventional hematological evaluation techniques, shortage of pathology laboratories, extreme workload and lack of trained or expert
hematopathologists results in wrong diagnosis.
Leukocytes (WBCs) are classified according to the
individuality and characteristics of their morphological structures, cytoplasm and nucleus . According to pathologists, white cells are classified into five categories
based on their characteristics and functions. 
\subsection{Literature Survey}

In digital image processing, the cell segmentation process is the separation of the cell from its complex background, and to break the cell into its morphological parts such as nucleus and
cytoplasm . Image pre-processing generally involves many different steps such as gray-scale image conversion. Initially, the white blood cell image taken from the microscope camera or collected by
lab is a color image. To make the image processing reliable for the classification, the original colored input images will be converted into gray level images. After conversion, contrast and brightness adjustment is applied to improve the quality of white cell image and improve the dynamic range of the image. Then histogram equalization is performed that adjusts intensity transformation, so that the histogram of the output image after brightness adjustment approximately matches a pre-defined histogram known as histogram specification. For extracting nucleus of leukocyte, global thresholding or any other detection method can be applied to the image. The morphological operations can be used to extract the white
blood cells (leukocytes) from other cells and background. Segmentation is a process of subdividing an image into the consequent parts or objects in the image, so each of these parts can be analyzed to extract some information which can be useful for high-level machine vision application. In case
of hematology, this information can be used for classification and identification of hematic disease.  \par
\subsection{Report overview}
The determination of the types of WBC like basophile,
eosinophils, lymphocyte and monocytes should be
done from the nucleus extraction. Since there are similar color scales in leukocyte with other blood components, for the purpose of reducing errors only the sub-images that contain the nucleus part are considered. Classification is the task to associate the appropriate class label (type of texture) with the blood test sample by using the measurements. Normally hematopathologists classify
leukocytes in five different classes i.e. lymphocytes, basophiles, neutrophils, monocytes and, eosinophils. They differ in their nucleus texture, nucleus size, affinity for different physiological colors, and functions for immune
system.
Feature selection is a very important step for classifying the white blood cells. According to hematopathologists, the white blood cell appears darker than the background, while red blood
cell (erythrocytes) appears in an intermediate intensity level. Platelets are much smaller than RBC and WBC. The chosen features affect the performance of classifiers.
The rest of the report is organized as: section III presents the preprocessing, section IV presents  feature extraction and section V discusses many classifiers used and finally conclusion is drawn comparing all the techniques over pipeline robustness, time and resource requirements in section V.
\begin{figure}
    \centering
    \includegraphics[scale=0.6]{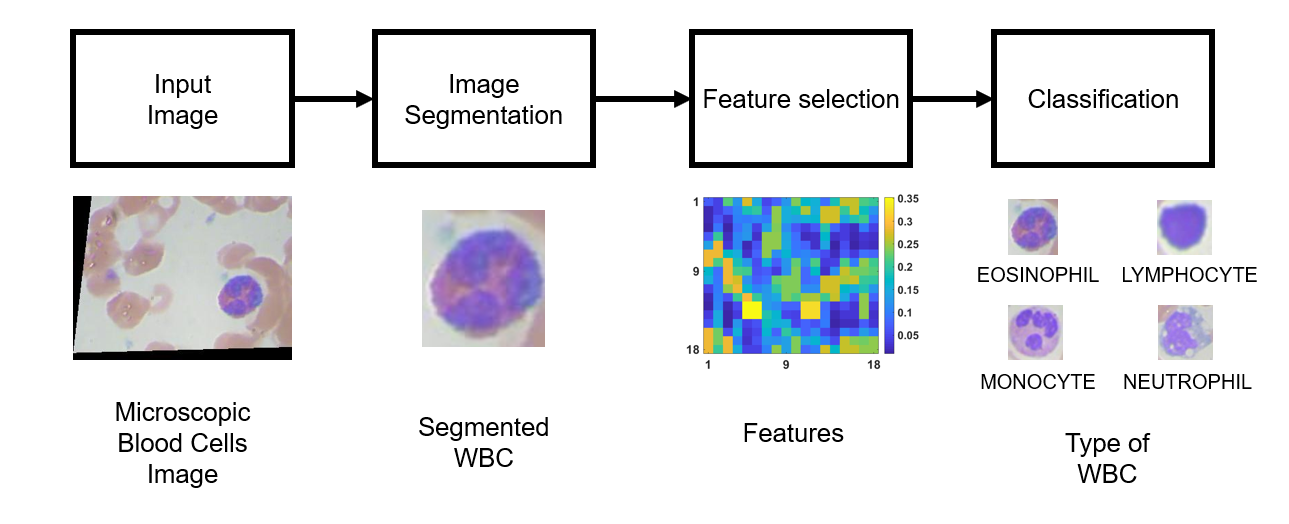}
    \caption{Overview Pipeline}
    \label{fig:overview}
\end{figure}
\subsection{Dataset}
For multi class cell detection we are using dataset from \url{https://www.kaggle.com/paultimothymooney/blood-cells}. This dataset contains 12,500 augmented images of blood cells (JPEG) with accompanying cell type labels (CSV). There are approximately 3,000 images for each of 4 different cell types grouped into 4 different folders (according to cell type). The cell types are Eosinophil, Lymphocyte, Monocyte, and Neutrophil. There are approximately 3,000 augmented images for each class of the 4 classes. The pre-augmented dataset contains 364 images across three classes: WBC (white blood cells), RBC (red blood cells), and Platelets. There are 4888 labels across 3 classes. The dataset is is made to a split of 80\% training images and 20\% Test images. All accuracy values to be reported are on Test dataset after training on Train dataset.   
\subsection{Manual Identification}

\begin{itemize}
\item \textbf{Limphocyte:} looks like well rounded purple colored potato. These are the easiest to identify.
\item \textbf{Monocyte:} The cell shape lookes roundish with skin-red color with some purple stuff inside. However the purple color is never fully covering the cell surface. Also, the purple colored portion of the cell is always in one continuous piece.
\item \textbf{Neutrophil:} The cell shape lookes roundish with skin-red color with some purple stuff inside. However the cell contains purple colored multiple whole-groundnuts inside it. The groundnuts could be disjointed within the cell.
\item \textbf{Eosinophil:} They look similar to Neutrophils. Which brings the importance of automated systems.
\end{itemize}
\section{Methods - Image Preprocessing}
In this section we will explore various pre-processing techniques and and narrow down to a pipeline for classification.
\subsection{Understanding the color space}
\begin{figure}
    \centering
    \includegraphics[scale=0.4]{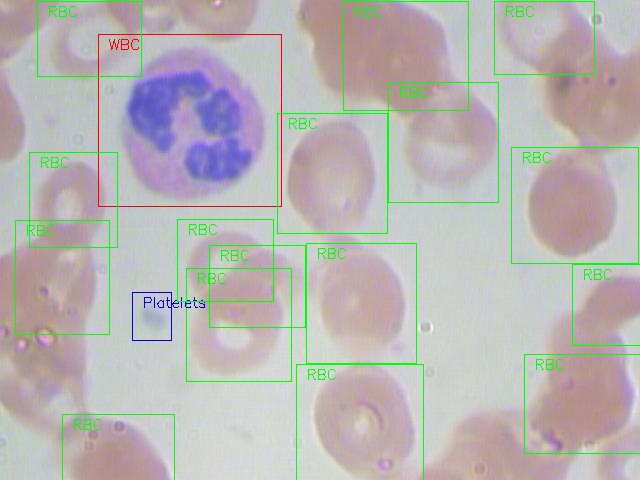}
    \caption{Sample annotated image to Red Blood Cells (RBC), White Blood cells (WBC), Blood Platlets}
    \label{fig:f1}
\end{figure}

\begin{figure}
    \centering
    \includegraphics[scale=0.4]{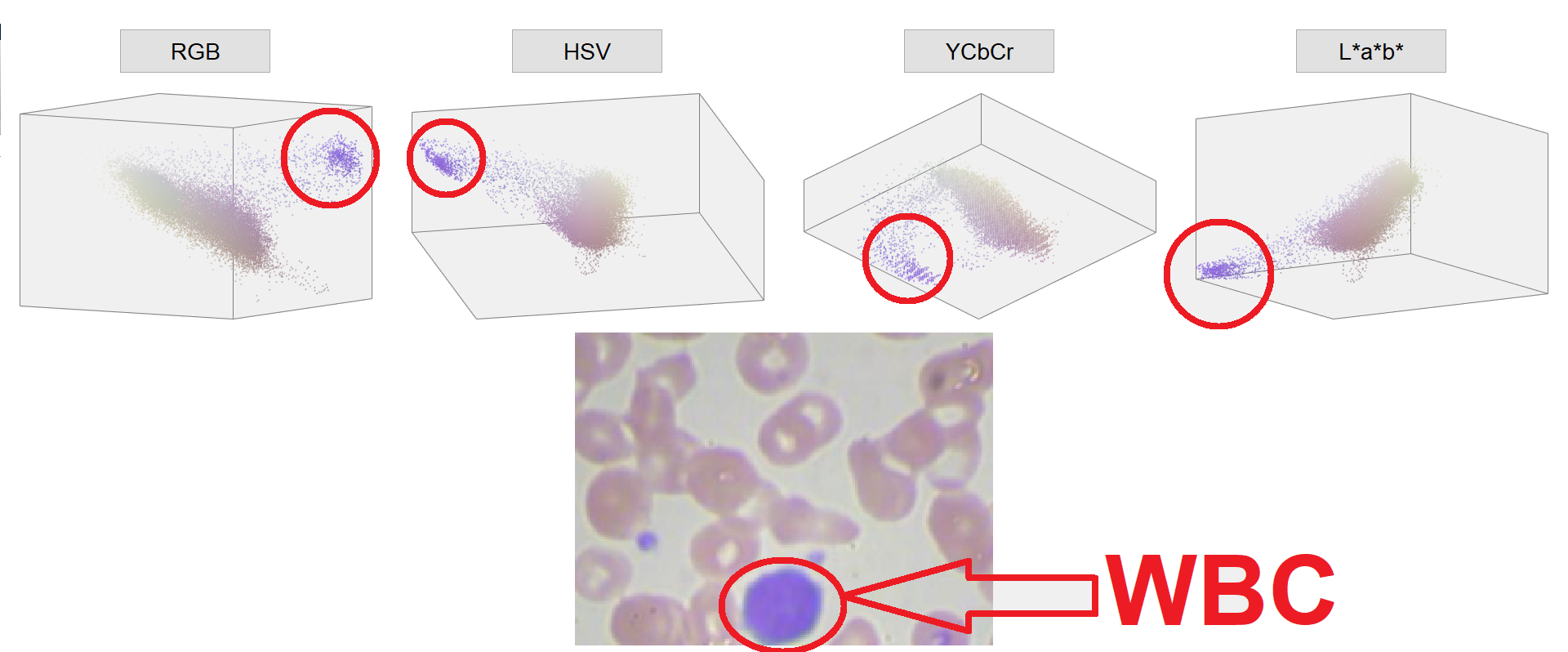}
    \caption{Above four subimages show decision space of below image in different color spaces (RGB, HSV, LAB, YCbCr). The red marked line shows the decision space of WBC color components}
    \label{fig:f2}
\end{figure}

Visually, white blood cells are easily distinguishable from the microscopic images as seen in figure \ref{fig:f1}. Digital images can be converted
to different computer graphics color spaces where there is a number of ways including
such as RGB (Red Green Blue),  HSV (Hue
Saturation and Value), YCbCr (Luma, blue-difference and red-difference chroma components), CIELAB color space.
To explore for image segmentation, a sample image is converted to different color spaces to find differentiable components in whole decision space. When color space is very distinguishable color decision boundaries can be drawn for image thresholding. As shown in figure \ref{fig:f2}, It is observed that in all color spaces the decision boundaries can be easily drawn.
\begin{table}[]
\caption{Comparison chart for WBC cell extraction over different color spaces}
\begin{tabular}{|c|c|c|c|c|c|}
\hline
\multicolumn{2}{|c|}{\textbf{Color Space}}                                                                          & \textbf{Min} & \textbf{Max} & \textbf{Color Space Conversion time}             & \textbf{IOU}                                   \\ \hline
                                                                                           & \textbf{R}             & 0            & 205          & \cellcolor[HTML]{34FF34}                         & \cellcolor[HTML]{34FF34}                       \\ \cline{2-4}
                                                                                           & \textbf{G}             & 0            & 190          & \cellcolor[HTML]{34FF34}                         & \cellcolor[HTML]{34FF34}                       \\ \cline{2-4}
\multirow{-3}{*}{\textbf{\begin{tabular}[c]{@{}c@{}}RGB \\ (UINT 8 Space)\end{tabular}}}   & \textbf{B}             & 195          & 255          & \multirow{-3}{*}{\cellcolor[HTML]{34FF34}None}   & \multirow{-3}{*}{\cellcolor[HTML]{34FF34}77\%} \\ \hline
                                                                                           & \textbf{H (degrees)}   & 113          & 273          & \cellcolor[HTML]{FCFF2F}                         & \cellcolor[HTML]{34FF34}                       \\ \cline{2-4}
                                                                                           & \textbf{S(Normalized)} & 0.136        & 1            & \cellcolor[HTML]{FCFF2F}                         & \cellcolor[HTML]{34FF34}                       \\ \cline{2-4}
\multirow{-3}{*}{\textbf{HSV}}                                                             & \textbf{V(Normalized)} & 0.615        & 1            & \multirow{-3}{*}{\cellcolor[HTML]{FCFF2F}Medium} & \multirow{-3}{*}{\cellcolor[HTML]{34FF34}78\%} \\ \hline
                                                                                           & \textbf{Y}             & 0            & 142          & \cellcolor[HTML]{67FD9A}                         & \cellcolor[HTML]{34FF34}                       \\ \cline{2-4}
                                                                                           & \textbf{Cb}            & 150          & 255          & \cellcolor[HTML]{67FD9A}                         & \cellcolor[HTML]{34FF34}                       \\ \cline{2-4}
\multirow{-3}{*}{\textbf{\begin{tabular}[c]{@{}c@{}}YCbCr \\ (UINT 8 Space)\end{tabular}}} & \textbf{Cr}            & 96           & 155          & \multirow{-3}{*}{\cellcolor[HTML]{67FD9A}Low}    & \multirow{-3}{*}{\cellcolor[HTML]{34FF34}76\%} \\ \hline
                                                                                           & \textbf{L}             & 48.915       & 56.259       & \cellcolor[HTML]{FE0000}                         & \cellcolor[HTML]{34FF34}                       \\ \cline{2-4}
                                                                                           & \textbf{A}             & -11.798      & 44.262       & \cellcolor[HTML]{FE0000}                         & \cellcolor[HTML]{34FF34}                       \\ \cline{2-4}
\multirow{-3}{*}{\textbf{LAB}}                                                             & \textbf{B}             & -61.422      & 18.364       & \multirow{-3}{*}{\cellcolor[HTML]{FE0000}High}   & \multirow{-3}{*}{\cellcolor[HTML]{34FF34}72\%} \\ \hline
\end{tabular}
\label{table1}
\end{table}
\par As shown in Table \ref{table1}, we have found out different threshold values for each channel over each space. The threshold values are chosen such that they have high precision over annotated dataset. These are qualitative values extracted over few images in each class. RGB data is existing color space of the dataset. For converting into YCbCr is also in UINT 8 space, the conversion computation is very fast. For conversion to LAB color space, the conversion time is very high due to floating point operations, same with HSV color space but comparative conversion time is lower. Over set of annotated images the selected threshold is done and it is observed that Intersection over Union observed are similar in all color spaces. So, If low computation time is required RGB can be used for further processing, The IOU extracted values are only qualitative as available dataset was object detection annotation but not semantic annotation. 

\subsection{Morphological operation exploration for removing other objects}
\begin{figure}
    \centering
    \includegraphics[scale=0.3]{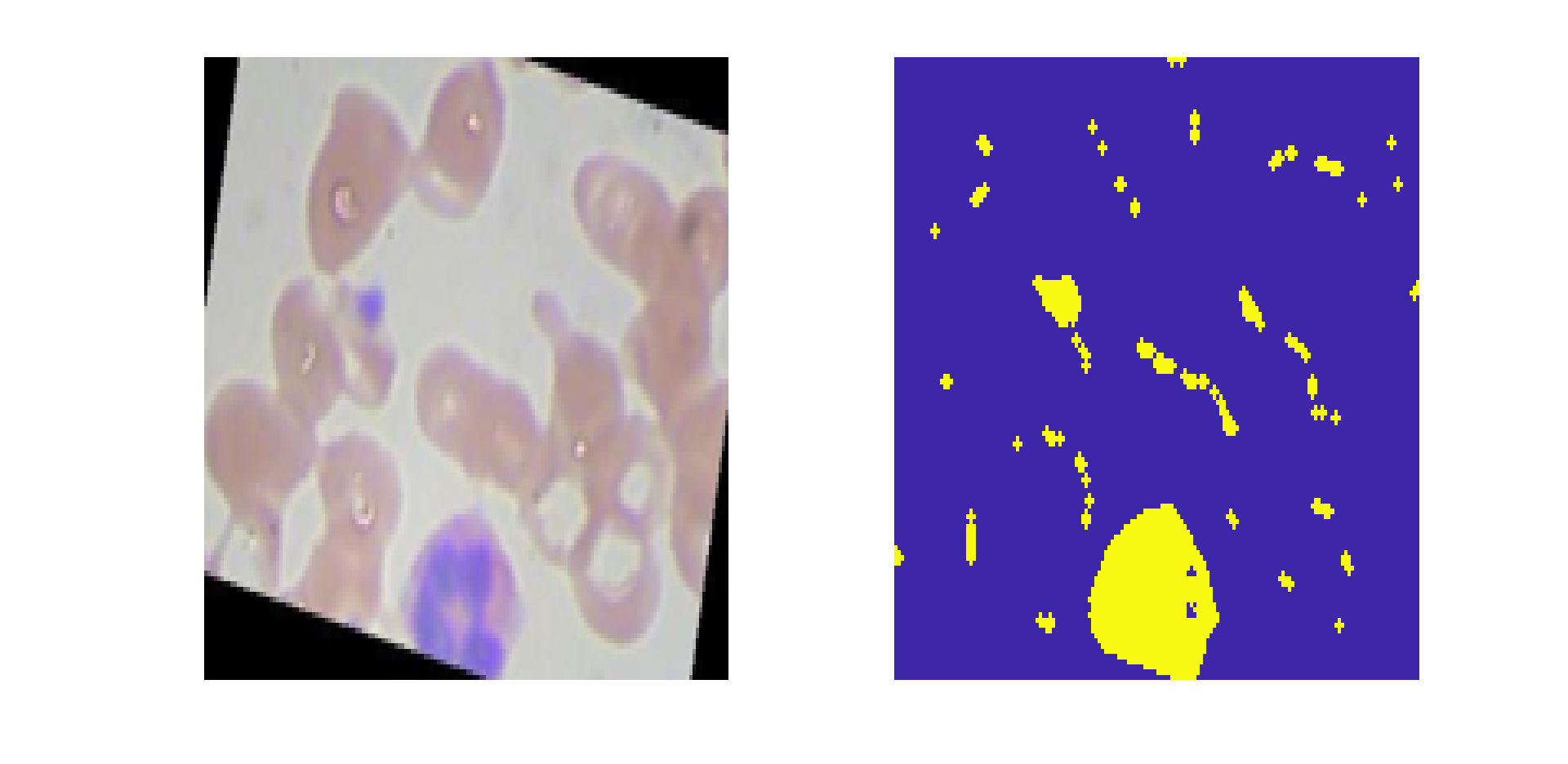}
    \caption{RGB based color threshold}
    \label{fig:f3}
\end{figure}
As shown in figure \ref{fig:f3}, there after color thresholding there are still few objects which were detected in color thresholding, the holes in the image must be filled and smaller blobs need to be eliminated.
\subsubsection{Dilation}: For filling holes image dilation can be used. In image dilation the basic effect of the operator/ structural element/ kernel on a binary image is to gradually enlarge the boundaries of regions of foreground pixels (i.e. yellow pixels in figure \ref{fig:f3}). Thus areas of foreground pixels grow in size while holes within those regions become smaller. As holes observed to be in shape of \textbf{disk} and size of \textbf{  20 to 40 sq.px}. So disk shaped kernel of radius \textbf{3} is used for dilation.Which fills the holes and gives extra boundary over blobs over binary mask image.

\subsubsection{Binary Connected Component Labelling}: As seen in figure \ref{fig:f3}, The major blob represents the WBC. We need to eliminate the smaller blobs. For this we need to label all the blobs to eliminate through thier region properties. Through Connected components labeling algorithms scans an image and groups its pixels into components based on pixel connectivity, i.e. all pixels in a connected component share similar pixel intensity values and are in some way connected with each other. Once all groups have been determined, each pixel is labeled with a graylevel or a color (color labeling) according to the component it was assigned to. Extracting and labeling of various disjoint and connected components in an image is central to many automated image analysis applications. The algorithm used is 

\begin{itemize}
\item If all four neighbors are 0, assign a new label to p, else
\item if only one neighbor has V={1}, assign its label to p, else
\item if more than one of the neighbors have V={1}, assign one of the labels to p and make a note of the equivalences.
\end{itemize}

For each label region properties are found out. The blob with largest area if selected eliminated which would corresponds to WBC. 

\subsubsection{Convex Hull} 
\begin{figure}
    \centering
    \includegraphics[scale=0.4]{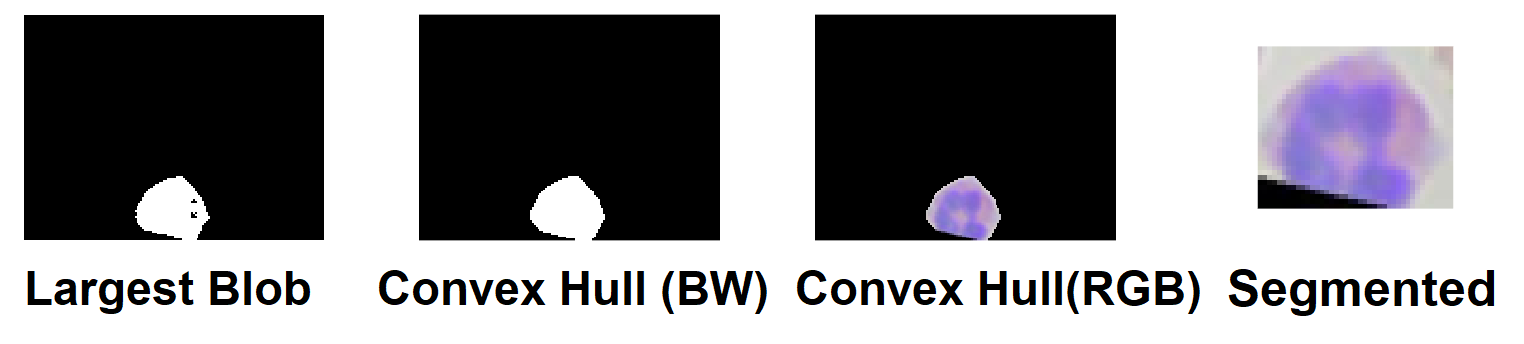}
    \caption{Largest blog to segmented image}
    \label{fig:f4}
\end{figure}

: As seen in figure \ref{fig:f4}. The convex hull of a binary image is the set of pixels included in the smallest convex polygon that surround all white pixels in the input.
\subsection{Exploring Geometrical Transformation/ Normalization}
As observed in the dataset the image size are of 320 X 240 pixels. And WBC nuclei is around 100 X 100 pixels. So, the extracted images are normalized to 128 X 128 pixels. For not distorting image more it is scaled it with bi-linear interpolation. I.e the output pixel value is a weighted average of pixels in the nearest 2-by-2 neighborhood.

\section{Methods Feature selection exploration}
\subsection{Feature extraction exploration}
\begin{figure}
    \centering
    \includegraphics[scale=1.1]{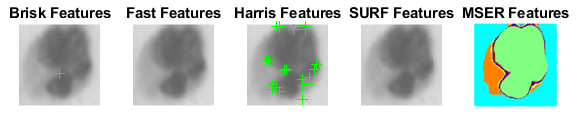}
    \caption{Local Features.}
    \label{fig:f5}
\end{figure}

\begin{figure}
    \centering
    \includegraphics[scale=1.1]{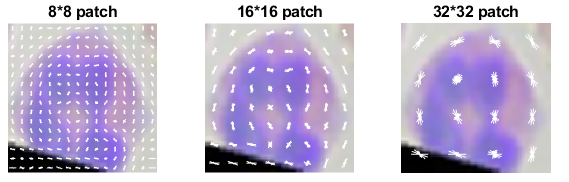}
    \caption{HOG features with different patch size.}
    \label{fig:f6}
\end{figure}

For each segmented image of 128 X 128 pixels. Only few features are discriminating features from one class to other. Local features and their descriptors, which are a compact vector representations of a local neighborhood, are the building blocks of many computer vision algorithms. Using local features enables these classification algorithms to better handle scale changes, rotation, and occlusion. Few features are
\begin{enumerate}
\item \textbf{For corner features} : Features from accelerated segment test (FAST), Harris-Stephens features, Oriented FAST and Rotated BRIEF (ORB), Binary Robust Invariant Scalable Keypoints (BRISK)
\item \textbf{For Blob Features} : Speeded-up robust features (SURF), KAZE, and Maximally stable external regions (MSER) features.
\item \textbf{Other features} : Histogram of oriented gradients (HOG) features.
\end{enumerate}
We can mix and match the detectors and the descriptors depending on the requirements of application.
On the segmented image we implemented various corner and blob features shown in figure \ref{fig:f5} and HOG features with different patch size \ref{fig:f6}.

\par For selecting the features extraction algorithm we have run the algorithm on all segmented images and quantified it through number of features extracted, homogeneity, number of distinctive features, time complexity. The obtained qualitative results are shown in Table \ref{t3}.

Extracted Harris features are observed to be have low features to be extracted, but might suffer in classification because observed low homogeneity and most components have high variance. Feature extraction become insignificant over augmented images. The observed detection time is high but can have low time complexity for feature matching as number of features are low.
Extracted FAST, BRISK, SURF features are observed to be suffer in classification because observed low homogeneity and most components have high variance, in few images there are low features detected and few it was not detected. Feature detection suffers drastically over augmented images. The features number can be enhanced by sharpening image, but it would be an additional step.
MSER based features is the most dense feature vector with over 150,000 points which is divided into 15 blobs. Being dense it is homogeneous and has low variance which tends to derive high time complexity for matching. But, detection time is qualitatively medium. 
\begin{table}[]
\caption{Comparison over various feature selection}
\label{t3}
\begin{tabular}{|c|c|c|c|c|c|c|c|c|}
\hline
\textbf{Algorithm}                                                                       & \textbf{\begin{tabular}[c]{@{}c@{}}Harris \\ features\end{tabular}} & \textbf{FAST}               & \textbf{BRISK}               & \textbf{SURF}               & \textbf{MSER}                & \textbf{\begin{tabular}[c]{@{}c@{}}HOG \\ {[}8,8{]}\end{tabular}} & \textbf{\begin{tabular}[c]{@{}c@{}}HOG \\ {[}16,16{]}\end{tabular}} & \textbf{\begin{tabular}[c]{@{}c@{}}HOG \\ {[}32,32{]}\end{tabular}} \\ \hline
\textbf{\begin{tabular}[c]{@{}c@{}}Total No.\\ of features\end{tabular}}                 & \cellcolor[HTML]{32CB00}180                                         & \cellcolor[HTML]{FE0000}-   & \cellcolor[HTML]{FE0000}-    & \cellcolor[HTML]{FE0000}-   & \cellcolor[HTML]{FE0000}150k & \cellcolor[HTML]{FFC702}8100                                      & \cellcolor[HTML]{FFC702}1764                                        & \cellcolor[HTML]{32CB00}324                                         \\ \hline
\textbf{Homogenity}                                                                      & \cellcolor[HTML]{FFC702}Low                                         & \cellcolor[HTML]{FE0000}No  & \cellcolor[HTML]{FE0000}No   & \cellcolor[HTML]{FE0000}No  & \cellcolor[HTML]{32CB00}Yes  & \cellcolor[HTML]{32CB00}Yes                                       & \cellcolor[HTML]{32CB00}Yes                                         & \cellcolor[HTML]{32CB00}Yes                                         \\ \hline
\textbf{\begin{tabular}[c]{@{}c@{}}Problem \\ with \\ Augumented \\ Images\end{tabular}} & \cellcolor[HTML]{FE0000}Yes                                         & \cellcolor[HTML]{FE0000}Yes & \cellcolor[HTML]{FE0000}Yes  & \cellcolor[HTML]{FE0000}Yes & \cellcolor[HTML]{32CB00}No   & \cellcolor[HTML]{32CB00}No                                        & \cellcolor[HTML]{32CB00}No                                          & \cellcolor[HTML]{32CB00}No                                          \\ \hline
\textbf{\begin{tabular}[c]{@{}c@{}}High \\ Variance\\ features (\%)\end{tabular}}        & \cellcolor[HTML]{FE0000}85                                          & \cellcolor[HTML]{FE0000}-   & \cellcolor[HTML]{FE0000}-    & \cellcolor[HTML]{FE0000}-   & \cellcolor[HTML]{FE0000}12   & \cellcolor[HTML]{FE0000}12                                        & \cellcolor[HTML]{32CB00}25                                          & \cellcolor[HTML]{32CB00}33                                          \\ \hline
\textbf{\begin{tabular}[c]{@{}c@{}}Color Space\\ (No. of\\ Channels)\end{tabular}}       & \cellcolor[HTML]{FFC702}1                                           & \cellcolor[HTML]{FFC702}1   & \cellcolor[HTML]{FFC702}1    & \cellcolor[HTML]{FFC702}1   & \cellcolor[HTML]{FFC702}1    & \cellcolor[HTML]{32CB00}3                                         & \cellcolor[HTML]{32CB00}3                                           & \cellcolor[HTML]{32CB00}3                                           \\ \hline
\textbf{\begin{tabular}[c]{@{}c@{}}Time\\  Complexity \\ (detection)\end{tabular}}       & \cellcolor[HTML]{FE0000}High                                        & \cellcolor[HTML]{32CB00}Low & \cellcolor[HTML]{FE0000}High & \cellcolor[HTML]{FFC702}Med & \cellcolor[HTML]{FFC702}Med  & \cellcolor[HTML]{32CB00}Low                                       & \cellcolor[HTML]{32CB00}Low                                         & \cellcolor[HTML]{32CB00}Low                                         \\ \hline
\textbf{\begin{tabular}[c]{@{}c@{}}Time \\ Complexity \\ (matching)\end{tabular}}        & \cellcolor[HTML]{32CB00}Low                                         & \cellcolor[HTML]{FE0000}-   & \cellcolor[HTML]{FE0000}-    & \cellcolor[HTML]{FE0000}-   & \cellcolor[HTML]{FE0000}High & \cellcolor[HTML]{FFC702}Med                                       & \cellcolor[HTML]{FFC702}Med                                         & \cellcolor[HTML]{32CB00}Low                                         \\ \hline
\end{tabular}
\end{table}
\par
 Histogram of Oriented Gradients (HOG) decomposes an image into small squared cells, computes an histogram of oriented gradients in each cell, normalizes the result using a block-wise pattern, and return a descriptor for each cell. The cell size selected with path size, we had explored 3 different patches 8 X 8, 16 X 16, 32 X 32 pixels. And finally we narrowed down to HOG with 32 X 32 pixels, as its observed to have low number of features and low time complexity for both matching and detection. It is observed to have high homogeneity and good number of components have high variance.

 \subsection{Methods - Dimensionality Reduction}
 Dimensionality reduction is the transformation of data from a high-dimensional space into a low-dimensional space so that the low-dimensional representation retains some meaningful properties of the original data, ideally close to its intrinsic dimension. Working in high-dimensional spaces can be undesirable for many reasons; raw data are often sparse as a consequence of the curse of dimensionality, and analyzing the data is usually computationally intractable. Feature selection approaches try to find a subset of the input variables (also called features or attributes). Classification can be done in the reduced space more accurately than in the original space.
 The main linear technique for dimensionality reduction, principal component analysis, performs a linear mapping of the data to a lower-dimensional space in such a way that the variance of the data in the low-dimensional representation is maximized. In practice, the covariance (and sometimes the correlation) matrix of the data is constructed and the eigenvectors on this matrix are computed. The eigenvectors that correspond to the largest eigenvalues (the principal components) can now be used to reconstruct a large fraction of the variance of the original data. Moreover, the first few eigenvectors can often be interpreted in terms of the large-scale physical behavior of the system, because they often contribute the vast majority of the system's energy, especially in low-dimensional systems. Still, this must be proven on a case-by-case basis as not all systems exhibit this behavior. The original space (with dimension of the number of points) has been reduced (with data loss, but hopefully retaining the most important variance) to the space spanned by a few eigenvectors.
 
\begin{figure}
     \centering
     \includegraphics[scale=0.5]{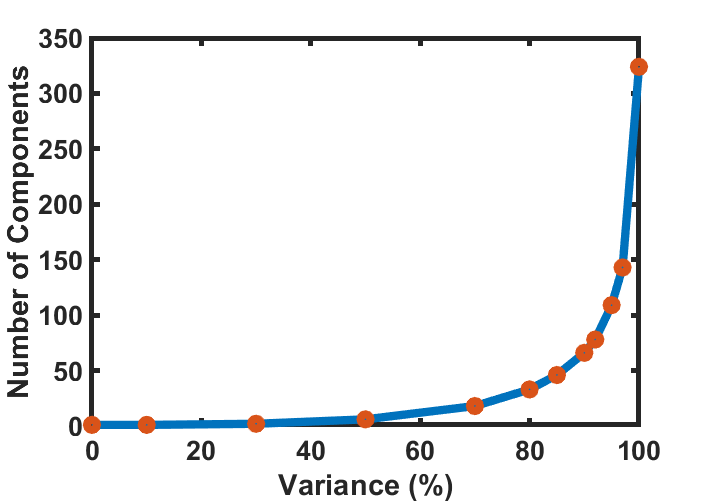}
     \caption{Variance vs Number of components (PCA)}
     \label{fig:f7}
 \end{figure}

 Using PCA when we observe the trend with variance over number of components is shown in figure \ref{fig:f7}.

\section{Classification Algorithm Exploration}
Image Classification is a fundamental task that attempts to comprehend an entire image as a whole. The goal is to classify the image by assigning it to a specific label. Typically, Image Classification refers to images in which only one object appears and is analyzed. In contrast, object detection involves both classification and localization tasks, and is used to analyze more realistic cases in which multiple objects may exist in an image.
For specific specific to our application of multi class cell classification, the machine learning algorithms need to classify the image to 4 different classes
\begin{enumerate}
\item Basophile,
\item Eosinophils, 
\item Lymphocyte and
\item Monocytes 
\end{enumerate}

\par So, we explore few Machine Learning algorithms for understanding the robustness, time and resource requirement.

\subsection{ML algorithms}
\textbf{Decsion trees :} Decision trees classify by sorting them down the tree from the root to some leaf node, with the leaf node providing the classification to the example. Each node in the tree acts as a test case for some attribute, and each edge descending from that node corresponds to one of the possible answers to the test case. This process is recursive in nature and is repeated for every sub tree rooted at the new nodes. We implemented decision trees with different splits, fine (100), medium  (20), Coarse (3). Fine Tree had given best accuracy, but took time in training, but very fast in prediction. It can be observed clearly that with PCA feature reduction helped in accuracy boost. Though, best test accuracy observed in decision trees is 69.8\%, which is very very low. Hence Decision trees can be used in preliminary classification as prediction speed is very high in resource constrained environments.

\textbf{Discriminant Analysis :} Given two or more groups or populations and a set of associated variables one often wants to locate a subset of the variables and associated functions of the subset that leads to maximum separation among the centroids of the groups. The exploratory multivariate procedure of determining variables and a reduced set of functions called discriminants or discriminant functions is called discriminant analysis. For implementation we have tried Linear Discriminant Analysis (LDA), Quadratic Discriminant Analysis (QDA).

It is observed that although prediction speed is moderate and training time is low, the accuracy of 85.6\% is not a reasonable accuracy. 

\begin{figure}
\centering  
\subfigure[Fine Tree]{\includegraphics[width=0.24\linewidth]{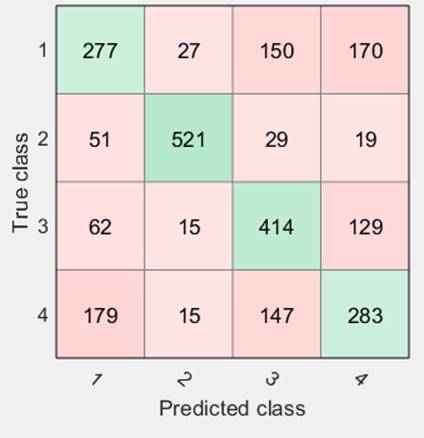}}
\subfigure[Medium Tree]{\includegraphics[width=0.24\linewidth]{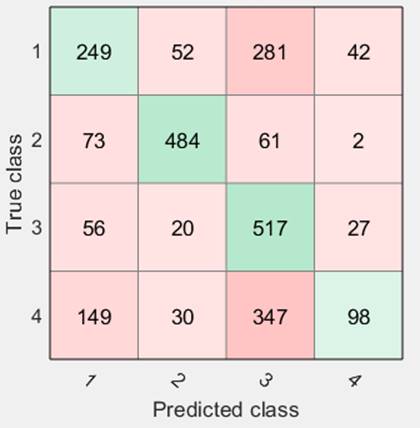}}
\subfigure[Coarse Tree]{\includegraphics[width=0.24\linewidth]{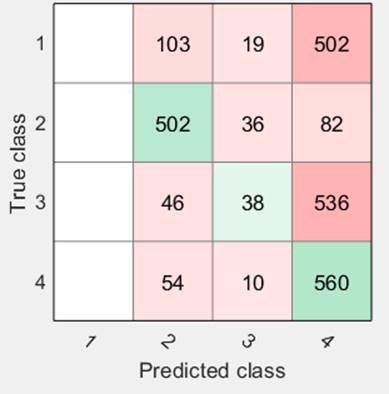}}
\subfigure[Linear Discriminant Analysis]{\includegraphics[width=0.24\linewidth]{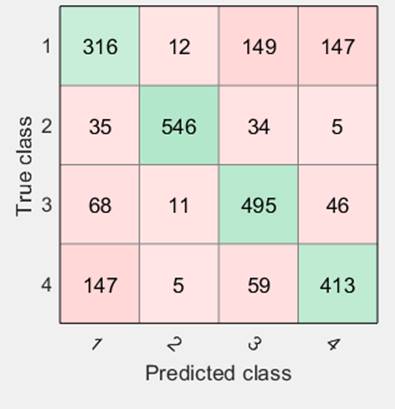}}
\subfigure[Quadratic Discriminant Analysis]{\includegraphics[width=0.24\linewidth]{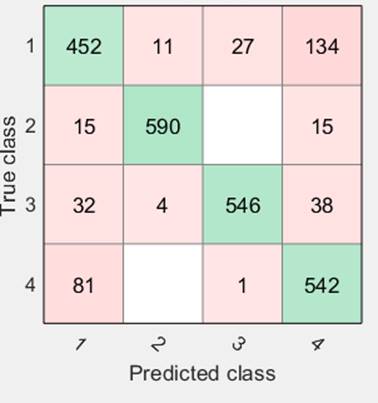}}
\subfigure[Naïve Bayes]{\includegraphics[width=0.24\linewidth]{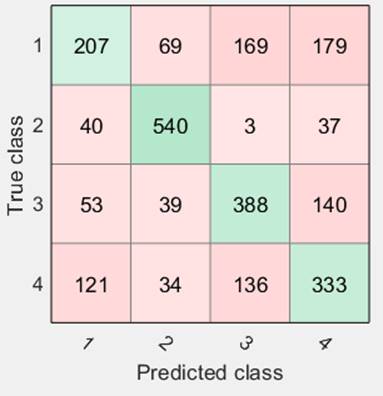}}
\subfigure[Linear SVM]{\includegraphics[width=0.24\linewidth]{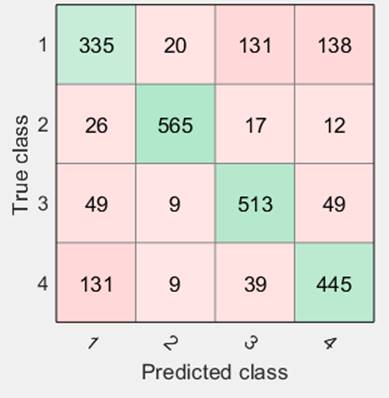}}
\subfigure[Quadratic SVM]{\includegraphics[width=0.24\linewidth]{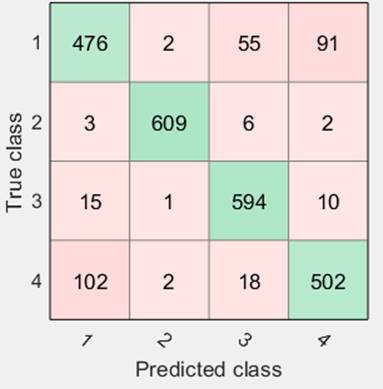}}
\subfigure[Cubic SVM]{\includegraphics[width=0.24\linewidth]{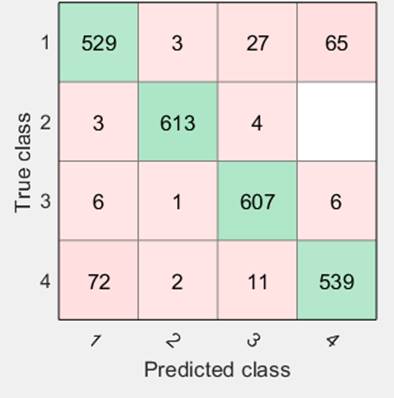}}
\subfigure[Gaussian SVM]{\includegraphics[width=0.24\linewidth]{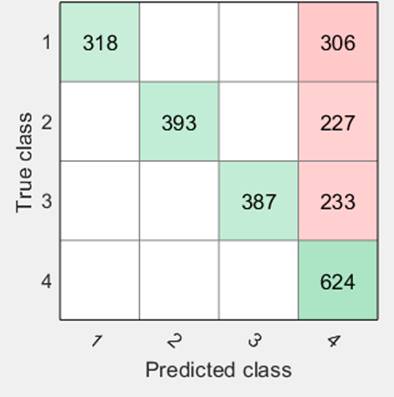}}
\subfigure[KNN (1)]{\includegraphics[width=0.24\linewidth]{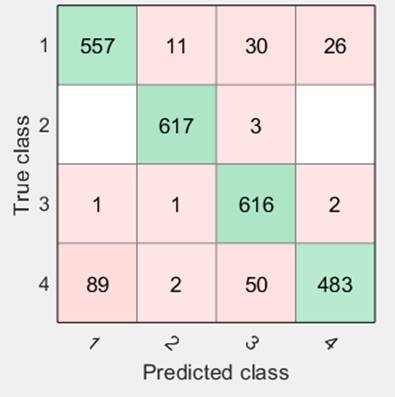}}
\subfigure[KNN (10)]{\includegraphics[width=0.24\linewidth]{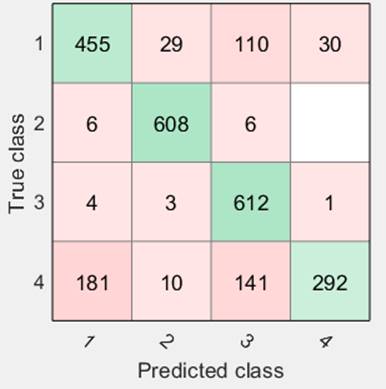}}
\caption{Confusion Matrix of various Machine Learning Algorithms without feature reduction}
\end{figure}

\textbf{Naive Bayes :} In statistics, Naive Bayes classifiers are a family of simple "probabilistic classifiers" based on applying Bayes' theorem with strong (naïve) independence assumptions between the features. They are among the simplest Bayesian network models. But they could be coupled with Kernel density estimation and achieve higher accuracy levels.

It is observed that accuracy is too low 61.5\% even with a PCA.

\textbf{Support Vector Machines :} A Support Vector Machine (SVM) is a discriminative classifier formally defined by a separating hyperplane. In other words, given labeled training data (supervised learning), the algorithm outputs an optimal hyperplane which categorizes new examples. In two dimensional space this hyperplane is a line dividing a plane in two parts where in each class lay in either side. We have tried to implement four different kernels, i.e Linear, Quadratic, Cubic, Gaussian.

Among all kernels Quadratic and Cubic kernels have shown highest accuracy i.e reasonable accuracy for our application. But training time is very high even after feature reduction, becoming the limitation for our application.

\textbf{K Nearest Neighbors :} The KNN algorithm is a robust and versatile classifier that is often used as a benchmark for more complex classifiers such as Artificial Neural Networks (ANN) and Support Vector Machines (SVM). Despite its simplicity, kNN can outperform more powerful classifiers,

Similarly kNN had shown highest accuracy i.e reasonable accuracy for our application. Training time being resonable. The prediction speed is also reasonable for resource constrained environments with PCA. This makes kNN (k=1), best in resource constrained environments.

\begin{figure}
\centering  
\subfigure[Fine Tree]{\includegraphics[width=0.24\linewidth]{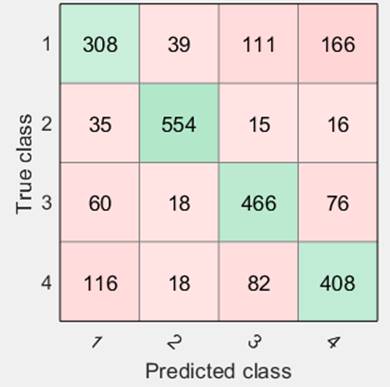}}
\subfigure[Medium Tree]{\includegraphics[width=0.24\linewidth]{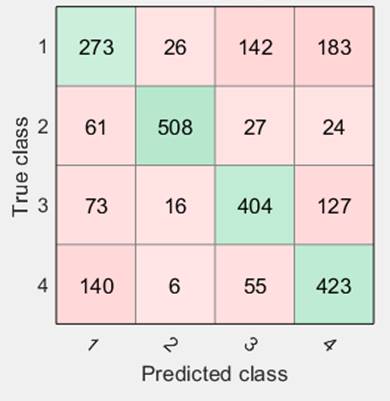}}
\subfigure[Coarse Tree]{\includegraphics[width=0.24\linewidth]{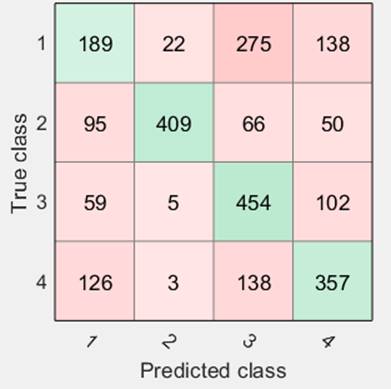}}
\subfigure[Linear Discriminant Analysis]{\includegraphics[width=0.24\linewidth]{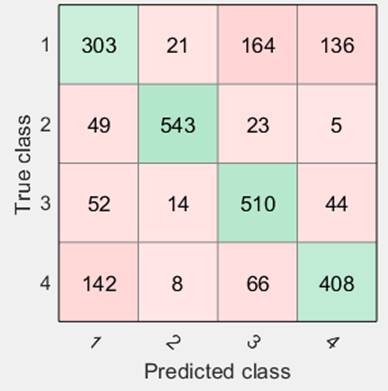}}
\subfigure[Quadratic Discriminant Analysis]{\includegraphics[width=0.24\linewidth]{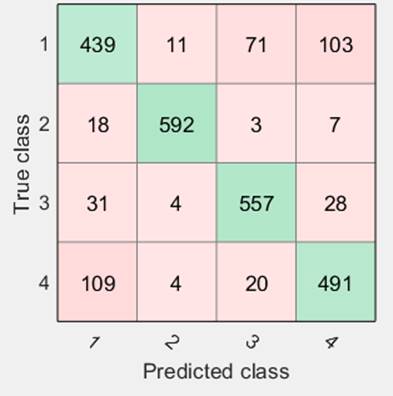}}
\subfigure[Naïve Bayes]{\includegraphics[width=0.24\linewidth]{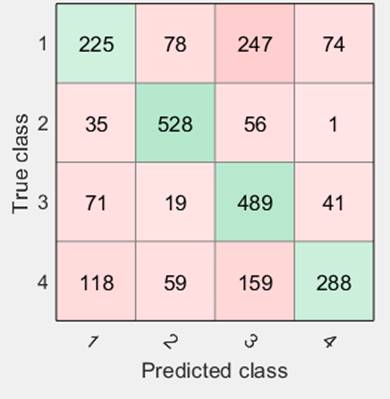}}
\subfigure[Linear SVM]{\includegraphics[width=0.24\linewidth]{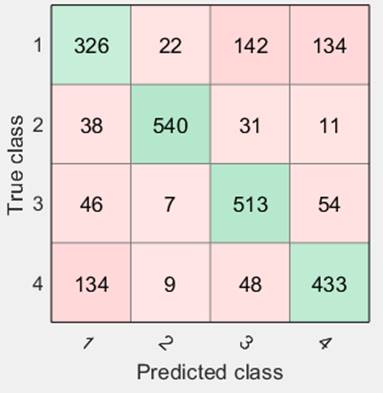}}
\subfigure[Quadratic SVM]{\includegraphics[width=0.24\linewidth]{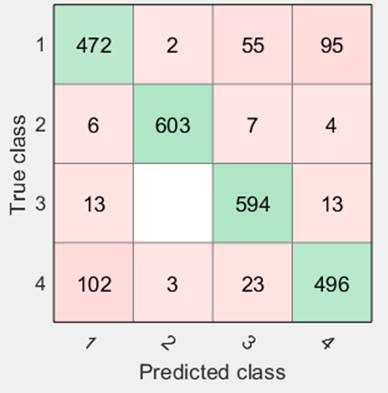}}
\subfigure[Cubic SVM]{\includegraphics[width=0.24\linewidth]{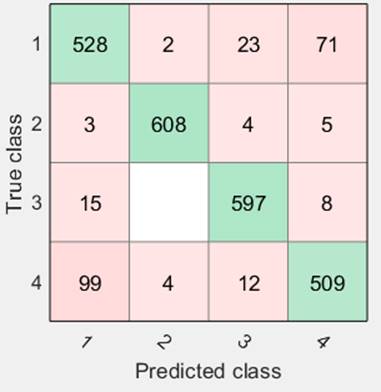}}
\subfigure[Gaussian SVM]{\includegraphics[width=0.24\linewidth]{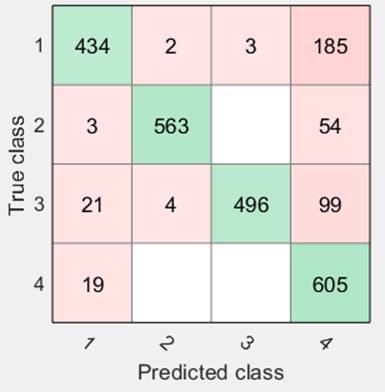}}
\subfigure[KNN (1)]{\includegraphics[width=0.24\linewidth]{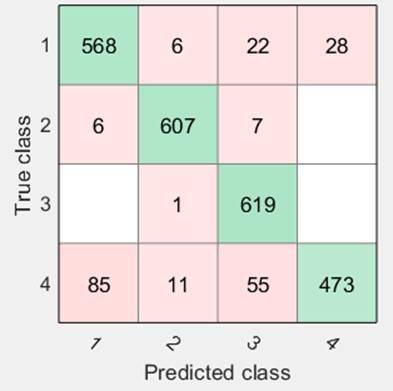}}
\subfigure[KNN (10)]{\includegraphics[width=0.24\linewidth]{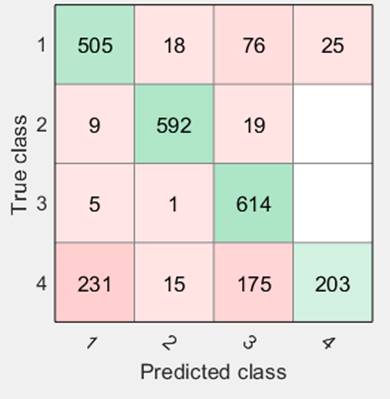}}
\caption{Confusion Matrix of various Machine Learning Algorithms with feature reduction (PCA var=95\%)}
\end{figure}

\begin{table}[]
\caption{Comparison table with various machine learning algorithms with respect to speed and robustness with and without PCA}
\centering
\begin{tabular}{|
>{\columncolor[HTML]{EFEFEF}}c |c|c|c|c|c|
>{\columncolor[HTML]{32CB00}}c |c|}
\hline
\cellcolor[HTML]{EFEFEF}                                                  & \cellcolor[HTML]{EFEFEF}                                     & \multicolumn{3}{c|}{\cellcolor[HTML]{C0C0C0}Without PCA}                                                                                                                                                                                                                                               & \multicolumn{3}{c|}{\cellcolor[HTML]{C0C0C0}With PCA}                                                                                                                                                                                                                                                  \\ \cline{3-8} 
\multirow{-2}{*}{\cellcolor[HTML]{EFEFEF}\textbf{Algorithm}}              & \multirow{-2}{*}{\cellcolor[HTML]{EFEFEF}\textbf{Specifics}} & \cellcolor[HTML]{EFEFEF}\textbf{\begin{tabular}[c]{@{}c@{}}Acc \\ (\%)\end{tabular}} & \cellcolor[HTML]{EFEFEF}\textbf{\begin{tabular}[c]{@{}c@{}}Prediction \\ Speed \\ (obs/sec)\end{tabular}} & \cellcolor[HTML]{EFEFEF}\textbf{\begin{tabular}[c]{@{}c@{}}Training \\ Time \\ (Secs)\end{tabular}} & \cellcolor[HTML]{EFEFEF}\textbf{\begin{tabular}[c]{@{}c@{}}Acc \\ (\%)\end{tabular}} & \cellcolor[HTML]{EFEFEF}\textbf{\begin{tabular}[c]{@{}c@{}}Prediction \\ Speed \\ (obs/sec)\end{tabular}} & \cellcolor[HTML]{EFEFEF}\textbf{\begin{tabular}[c]{@{}c@{}}Training \\ Time \\ (Secs)\end{tabular}} \\ \hline
\textbf{Tree}                                                             & Fine(100)                                                    & \cellcolor[HTML]{FE0000}60.1                                                         & \cellcolor[HTML]{32CB00}50000                                                                             & \cellcolor[HTML]{FFCB2F}7.9454                                                                      & \cellcolor[HTML]{FE0000}69.8                                                         & 15000                                                                                                     & \cellcolor[HTML]{FFCB2F}5.9236                                                                      \\ \hline
\textbf{Tree}                                                             & Medium (20)                                                  & \cellcolor[HTML]{FE0000}54.2                                                         & \cellcolor[HTML]{32CB00}53000                                                                             & \cellcolor[HTML]{FFCB2F}4.9511                                                                      & \cellcolor[HTML]{FE0000}64.6                                                         & 16000                                                                                                     & \cellcolor[HTML]{32CB00}3.3087                                                                      \\ \hline
\textbf{Tree}                                                             & Coarse (3)                                                   & \cellcolor[HTML]{FE0000}44.2                                                         & \cellcolor[HTML]{32CB00}49000                                                                             & \cellcolor[HTML]{32CB00}3.0678                                                                      & \cellcolor[HTML]{FE0000}56.6                                                         & 17000                                                                                                     & \cellcolor[HTML]{32CB00}2.8411                                                                      \\ \hline
\textbf{\begin{tabular}[c]{@{}c@{}}Discriminant \\ Analysis\end{tabular}} & Linear                                                       & \cellcolor[HTML]{FFC702}71.2                                                         & \cellcolor[HTML]{FFC702}21000                                                                             & \cellcolor[HTML]{32CB00}1.4384                                                                      & \cellcolor[HTML]{FFC702}70.9                                                         & 14000                                                                                                     & \cellcolor[HTML]{32CB00}2.9011                                                                      \\ \hline
\textbf{\begin{tabular}[c]{@{}c@{}}Discriminant \\ Analysis\end{tabular}} & Quadratic                                                    & \cellcolor[HTML]{32CB00}85.6                                                         & \cellcolor[HTML]{FFC702}21000                                                                             & \cellcolor[HTML]{32CB00}1.5984                                                                      & \cellcolor[HTML]{32CB00}83.6                                                         & 14000                                                                                                     & \cellcolor[HTML]{32CB00}2.1812                                                                      \\ \hline
\textbf{Naïve Bayes}                                                      & -                                                            & \cellcolor[HTML]{FE0000}59                                                           & \cellcolor[HTML]{FFC702}28000                                                                             & \cellcolor[HTML]{32CB00}1.782                                                                       & \cellcolor[HTML]{FE0000}61.5                                                         & 17000                                                                                                     & \cellcolor[HTML]{32CB00}2.8145                                                                      \\ \hline
\textbf{SVM}                                                              & Linear                                                       & \cellcolor[HTML]{FFC702}74.7                                                         & \cellcolor[HTML]{FFC702}21000                                                                             & \cellcolor[HTML]{FE0000}38.866                                                                      & \cellcolor[HTML]{FFC702}72.8                                                         & 13000                                                                                                     & \cellcolor[HTML]{FE0000}30.133                                                                      \\ \hline
\textbf{SVM}                                                              & Quadratic                                                    & \cellcolor[HTML]{32CB00}87.7                                                         & \cellcolor[HTML]{FE0000}1800                                                                              & \cellcolor[HTML]{FE0000}49.651                                                                      & \cellcolor[HTML]{32CB00}87.7                                                         & \cellcolor[HTML]{FE0000}2200                                                                              & \cellcolor[HTML]{FE0000}54.421                                                                      \\ \hline
\textbf{SVM}                                                              & Cubic                                                        & \cellcolor[HTML]{32CB00}92                                                           & \cellcolor[HTML]{FE0000}1900                                                                              & \cellcolor[HTML]{FE0000}55.358                                                                      & \cellcolor[HTML]{32CB00}90.1                                                         & \cellcolor[HTML]{FE0000}4000                                                                              & \cellcolor[HTML]{FE0000}44.66                                                                       \\ \hline
\textbf{SVM}                                                              & Gaussian                                                     & \cellcolor[HTML]{FE0000}69.2                                                         & \cellcolor[HTML]{FE0000}270                                                                               & \cellcolor[HTML]{FE0000}118.34                                                                      & \cellcolor[HTML]{32CB00}84.3                                                         & \cellcolor[HTML]{FE0000}1400                                                                              & \cellcolor[HTML]{FE0000}55.827                                                                      \\ \hline
\textbf{KNN}                                                              & k=1                                                          & \cellcolor[HTML]{32CB00}91.4                                                         & \cellcolor[HTML]{FE0000}580                                                                               & \cellcolor[HTML]{FFCB2F}9.147                                                                       & \cellcolor[HTML]{32CB00}91.1                                                         & \cellcolor[HTML]{FFC702}1700                                                                              & \cellcolor[HTML]{32CB00}4.4275                                                                      \\ \hline
\textbf{KNN}                                                              & k=10                                                         & \cellcolor[HTML]{FFC702}79.1                                                         & \cellcolor[HTML]{FE0000}550                                                                               & \cellcolor[HTML]{FFCB2F}9.558                                                                       & \cellcolor[HTML]{FFC702}76.9                                                         & \cellcolor[HTML]{FFC702}1700                                                                              & \cellcolor[HTML]{32CB00}4.3354                                                                      \\ \hline
\end{tabular}
\end{table}
\subsection{Deep Neural Networks exploration}
A deep neural network (DNN) is an artificial neural network (ANN) with multiple layers between the input and output layers. The DNN finds the correct mathematical manipulation to turn the input into the output, whether it be a linear relationship or a non-linear relationship. The network moves through the layers calculating the probability of each output. For example, a DNN that is trained to recognize dog breeds will go over the given image and calculate the probability that the dog in the image is a certain breed. The user can review the results and select which probabilities the network should display (above a certain threshold, etc.) and return the proposed label. Each mathematical manipulation as such is considered a layer, and complex DNN have many layers, hence the name "deep" networks.

DNNs can model complex non-linear relationships. DNN architectures generate compositional models where the object is expressed as a layered composition of primitives. The extra layers enable composition of features from lower layers, potentially modeling complex data with fewer units than a similarly performing shallow network.

Deep architectures include many variants of a few basic approaches. Each architecture has found success in specific domains. It is not always possible to compare the performance of multiple architectures, unless they have been evaluated on the same data sets.

DNNs are typically feedforward networks in which data flows from the input layer to the output layer without looping back. At first, the DNN creates a map of virtual neurons and assigns random numerical values, or "weights", to connections between them. The weights and inputs are multiplied and return an output between 0 and 1. If the network did not accurately recognize a particular pattern, an algorithm would adjust the weights. That way the algorithm can make certain parameters more influential, until it determines the correct mathematical manipulation to fully process the data.

Though DNN had shown good classification accuracy and easy to implement nature for real time application, but lacks explainability. We implemented few DNN architectures ranging from from low memory foot print (MobilenetV2, DenseNet121, NASNetMobile), medium memory foot print (InceptionV3,ResNet50V2), large memory foot print (VGG16). We tried to implement architectures in low resource environments like CPU (Computer Processing Unit) and high resource environments like GPU (Graphical Processing Unit).

\begin{table}[]
\caption{Comparison table with various Deep Neural Network architectures with respect to speed and robustness with and without preprocessing}
\centering
\begin{tabular}{|l|l|l|
>{\columncolor[HTML]{FE0000}}l |l|l|l|l|}
\hline
\cellcolor[HTML]{C0C0C0}{\color[HTML]{333333} }                                 & \cellcolor[HTML]{C0C0C0}{\color[HTML]{333333} }                                                                                  & \multicolumn{6}{l|}{\cellcolor[HTML]{C0C0C0}{\color[HTML]{333333} \textbf{Training Time (min)}}}                                                                                                                                                                                                                                                                                            \\ \cline{3-8} 
\cellcolor[HTML]{C0C0C0}{\color[HTML]{333333} }                                 & \cellcolor[HTML]{C0C0C0}{\color[HTML]{333333} }                                                                                  & \multicolumn{3}{l|}{\cellcolor[HTML]{C0C0C0}{\color[HTML]{333333} \textbf{Without   Preprocessing}}}                                                                                         & \multicolumn{3}{l|}{\cellcolor[HTML]{C0C0C0}{\color[HTML]{333333} \textbf{With   Preprocessing}}}                                                                                            \\ \cline{3-8} 
\multirow{-3}{*}{\cellcolor[HTML]{C0C0C0}{\color[HTML]{333333} \textbf{Model}}} & \multirow{-3}{*}{\cellcolor[HTML]{C0C0C0}{\color[HTML]{333333} \textbf{\begin{tabular}[c]{@{}l@{}}Weight \\ Size\end{tabular}}}} & \cellcolor[HTML]{C0C0C0}{\color[HTML]{333333} \textbf{Accuracy}} & \cellcolor[HTML]{C0C0C0}{\color[HTML]{333333} \textbf{CPU}} & \cellcolor[HTML]{C0C0C0}{\color[HTML]{333333} \textbf{GPU}} & \cellcolor[HTML]{C0C0C0}{\color[HTML]{333333} \textbf{Accuracy}} & \cellcolor[HTML]{C0C0C0}{\color[HTML]{333333} \textbf{CPU}} & \cellcolor[HTML]{C0C0C0}{\color[HTML]{333333} \textbf{GPU}} \\ \hline
{\color[HTML]{333333} \textbf{MobilenetV2}}                                     & {\color[HTML]{333333} Low}                                                                                                       & \cellcolor[HTML]{FFC702}{\color[HTML]{333333} 87.4}              & {\color[HTML]{333333} 84.2}                                 & \cellcolor[HTML]{32CB00}{\color[HTML]{333333} 6.4}          & \cellcolor[HTML]{FFC702}{\color[HTML]{333333} 88.42}             & \cellcolor[HTML]{32CB00}{\color[HTML]{333333} 39}           & \cellcolor[HTML]{32CB00}{\color[HTML]{333333} 3.4}          \\ \hline
{\color[HTML]{333333} \textbf{DenseNet121}}                                     & {\color[HTML]{333333} Low}                                                                                                       & \cellcolor[HTML]{FFC702}{\color[HTML]{333333} 87.8}              & {\color[HTML]{333333} 144}                                  & \cellcolor[HTML]{FFC702}{\color[HTML]{333333} 12.6}         & \cellcolor[HTML]{FFC702}{\color[HTML]{333333} 89}                & \cellcolor[HTML]{32CB00}{\color[HTML]{333333} 70}           & \cellcolor[HTML]{FFC702}{\color[HTML]{333333} 6}            \\ \hline
{\color[HTML]{333333} \textbf{NASNetMobile}}                                    & {\color[HTML]{333333} Low}                                                                                                       & \cellcolor[HTML]{FE0000}{\color[HTML]{333333} 72}                & {\color[HTML]{333333} 277.8}                                & \cellcolor[HTML]{32CB00}{\color[HTML]{333333} 30.4}         & \cellcolor[HTML]{FE0000}{\color[HTML]{333333} 78}                & \cellcolor[HTML]{FE0000}{\color[HTML]{333333} 143}          & \cellcolor[HTML]{32CB00}{\color[HTML]{333333} 12}           \\ \hline
{\color[HTML]{333333} \textbf{InceptionV3}}                                     & {\color[HTML]{333333} Med}                                                                                                       & \cellcolor[HTML]{32CB00}{\color[HTML]{333333} 90.2}              & {\color[HTML]{333333} 261}                                  & \cellcolor[HTML]{32CB00}{\color[HTML]{333333} 27.6}         & \cellcolor[HTML]{32CB00}{\color[HTML]{333333} 92}                & \cellcolor[HTML]{FE0000}{\color[HTML]{333333} 142}          & \cellcolor[HTML]{32CB00}{\color[HTML]{333333} 13}           \\ \hline
{\color[HTML]{333333} \textbf{ResNet50V2}}                                      & {\color[HTML]{333333} Med}                                                                                                       & \cellcolor[HTML]{FFC702}{\color[HTML]{333333} 89.3}              & {\color[HTML]{333333} 136}                                  & \cellcolor[HTML]{FFC702}{\color[HTML]{333333} 16.4}         & \cellcolor[HTML]{32CB00}{\color[HTML]{333333} 91}                & \cellcolor[HTML]{FE0000}{\color[HTML]{333333} 75}           & \cellcolor[HTML]{FFC702}{\color[HTML]{333333} 7}            \\ \hline
{\color[HTML]{333333} \textbf{VGG16}}                                           & {\color[HTML]{333333} High}                                                                                                      & \cellcolor[HTML]{FFC702}{\color[HTML]{333333} 89}                & {\color[HTML]{333333} 159.6}                                & \cellcolor[HTML]{FFC702}{\color[HTML]{333333} 18.6}         & \cellcolor[HTML]{FFC702}{\color[HTML]{333333} 86}                & \cellcolor[HTML]{FE0000}{\color[HTML]{333333} 84}           & \cellcolor[HTML]{FFC702}{\color[HTML]{333333} 9}            \\ \hline
\end{tabular}
\end{table}

\section{Conclusion}
We tried to conclude by giving few insights from our research specific to the multi class blood cell classification application. The problem can be solved in RGB colour space, being robust and Resource efficient. The shapes of white blood cells becoming key differentiating feature, HOG features can be chosen as key feature vector. To make it resource efficient PCA up to var 95\% can be used. Among all ML algorithms, KNN was showing highest accuracy, with PCA it also shows a speedup for prediction. With or Without pre-processing DNN had shown good accuracy values, although with pre-processing DNN had made DNNs speedup upto 50\% and a marginal change of accuracy. With No brainer, negating the pre-processing MobilenetV2 can be used in resource constrained environment. And InceptionV3 is very robust in unconstrained environment. Although with pre-processing pipeline KNN showed best results being robust with 91.1\% accuracy and very very low inference and training times even in a resource constrained environment. This can be further extende by using multiple algorithms in different levels, boosting the accuracy, Hence, This was a conclusive exploration over multiple algorithms for solving the muli class cell classification.

\end{document}